\documentclass[letterpaper, 10 pt, conference]{ieeeconf}  
\usepackage{amssymb}

\let\labelindent\undefined

\usepackage{multicol}
\usepackage{tikz}
\usepackage{graphicx}
\usepackage{booktabs}
\usepackage{amsmath}
\usepackage[dvipsnames]{xcolor}
\usepackage{listings}
\usepackage[ruled,linesnumberedhidden,vlined]{algorithm2e}

\usepackage{comment}
\usepackage{wrapfig}
\usepackage{cuted}
\usepackage[hidelinks]{hyperref}
\usepackage[capitalise]{cleveref}
\usepackage[shortlabels]{enumitem}
\usepackage[acronym]{glossaries}
\usepackage{glossaries-extra}
\usepackage{authblk}

\Crefname{section}{Sec.}{Secs.}

\newenvironment{tightitemize}[1][$\bullet$]{
\begin{list}{#1}{
    \setlength{\leftmargin}{0pt}
    \setlength{\itemindent}{1.2em} 
    \setlength{\labelwidth}{1.2em}
    \setlength{\labelsep}{0.5em}
    \setlength{\itemsep}{0pt}
    \setlength{\parsep}{0pt}
    \setlength{\topsep}{0pt}
}}{\end{list}}

\glsdisablehyper
\newcommand\myparagraph[1]{\noindent{}\textbf{#1}\quad{}}
\setabbreviationstyle[acronym]{long-short}
\setabbreviationstyle[short]{short-nolong}
\newacronym[category={acronym}]{pddl}{\textsc{pddl}}{Planning Domain Definition Language}
\newacronym[category={short}]{mdp}{\textsc{mdp}}{Markov Decision Process}
\newacronym[category={short}]{pomdp}{\textsc{pomdp}}{Partially Observable Markov Decision Process}
\newacronym[category={short}]{lomdp}{\textsc{lomdp}}{Locally Observable Markov Decision Process}
\newacronym[category={acronym}]{LOMDP}{\textsc{lomdp}}{\emph{locally observable} \textsc{mdp}}
\newacronym[category={acronym}]{vlm}{\textsc{vlm}}{Vision Language Models}
\newacronym[category={acronym}]{lsp}{\textsc{lsp}}{Learning over Subgoal Planning}
\newacronym[category={short}]{fastdownward}{\textsc{FastDownward}}{FastDownward}
\newacronym[category={short}]{procthor}{\textsc{ProcTHOR}}{ProcTHOR}
\newacronym[category={short}]{find}{\texttt{find}}{find}
\newacronym[category={acronym}]{LIOS}{\textsc{lios}}{learning-informed object search}

\definecolor{hlcolor}{HTML}{E8E8FF}
\newcommand{\boldtt}[1]{%
    \texttt{#1}%
}

\newcommand{\Ounsearched}{\mathcal{O}_{\textsc{\tiny{}unsearched}}}
\newcommand{\Omissing}{\mathcal{O}_{\textsc{\tiny{}miss}}}
\newcommand{\Oknown}{\mathcal{O}_{\textsc{\tiny{}known}}}
\newcommand{\Pfound}{P_{\textsc{\tiny{}found}}}

\newcommand*\qnum[1]{\tikz[baseline=(char.base)]{            \node[shape=circle,draw,inner sep=1pt] (char) {\small{}#1};}}

\newif\ifreview

\ifreview
  \newcommand{\rev}[1]{\textcolor{blue}{#1}}
\else
  \newcommand{\rev}[1]{#1}
\fi

\title{\LARGE \bf
Effective Task Planning with Missing Objects using Learning-Informed Object Search
}

\author{
    Raihan Islam Arnob\textsuperscript{*}, 
    Max Merlin\textsuperscript{\dag}, 
    Abhishek Paudel\textsuperscript{*}, 
    Benned Hedegaard\textsuperscript{\dag}, \\
    George Konidaris\textsuperscript{\dag}, 
    and Gregory J. Stein\textsuperscript{*}
}

\affil{
    \textsuperscript{*}George Mason University, 
    \textsuperscript{\dag}Brown University
}

\begin{document}

\maketitle
\thispagestyle{empty}
\pagestyle{empty}


\begin{abstract}
Task planning for mobile robots often assumes full environment knowledge and so popular approaches, like planning via the \gls{pddl}, cannot plan when the locations of task-critical objects are unknown.
Recent learning-driven object search approaches are effective, but operate as standalone tools and so are not straightforwardly incorporated into full task planners, which must additionally determine both what objects are necessary and when in the plan they should be sought out.
To address this limitation, we develop a planning framework centered around novel model-based \gls{LIOS} actions: each a policy that aims to find and retrieve a single object.
High-level planning treats \gls{LIOS} actions as deterministic and so---informed by model-based calculations of the expected cost of each---generates plans that interleave search and execution for effective, sound, and complete learning-informed task planning despite uncertainty. 
Our work effectively reasons about uncertainty while maintaining compatibility with existing full-knowledge solvers.
In simulated \gls{procthor} homes \rev{and in the real world}, our approach outperforms non-learned and learned baselines on tasks including retrieval and meal prep.
\end{abstract}


\section{Introduction}
\label{sec:intro}

Task planning approaches have evolved significantly over the past five decades. Much of this progress is owed to the development of flexible languages for specifying planning problems (e.g., \gls{pddl}~\cite{aeronautiques1998pddl, fox2003pddl2}) and powerful solvers (e.g., the popular \gls{fastdownward}~\cite{helmert2006fast}), which have enabled successful and effective task completion from tabletop scenarios to large home-scale spaces. However, these tools are not straightforwardly applicable when the environment is not fully known and task-relevant objects are missing, an unavoidable scenario for mobile robots operating in the wild. Robots must \emph{search} for missing objects to complete tasks, a planning scenario made challenging by both a lack of direct knowledge about where missing objects may be and the computational burden of planning with search actions, which are inherently non-deterministic.

The recent \gls{LOMDP} model~\cite{merlin2024robot} provides a framework for leveraging full-knowledge planners despite partial observability by separating known-space actions from those that search: actions that incrementally expand the robot's \emph{envelope of observability} until the plan can be completed with the set of discovered objects.
However, effective behavior in a \gls{LOMDP} setting requires careful consideration of \emph{how} the robot should best expand this envelope of observability to quickly complete the task, raising key questions the planner must address: \qnum{1} how to quickly find missing objects, \qnum{2} what objects should be sought out, and \qnum{3} when during execution should the robot search.
These three questions comprise the object scouting problem~\cite{spop}.
Recent developments leveraging machine learning for cost-effective object search~\cite{chaplot2020object, krantz2023navigating, chang2023goat} provide a potential answer to \qnum{1}, as they have proven effective tools for quickly locating unseen objects in previously-unseen environments. 
However, such approaches are developed as standalone modules and so it is unclear how to to integrate them with a planner in ways that would address \qnum{2} and \qnum{3}---especially when there is considerable flexibility in how a task could be completed.

\begin{figure*}[t]
  \centering
  \vspace{.5em}
  \includegraphics[width=0.96\textwidth]{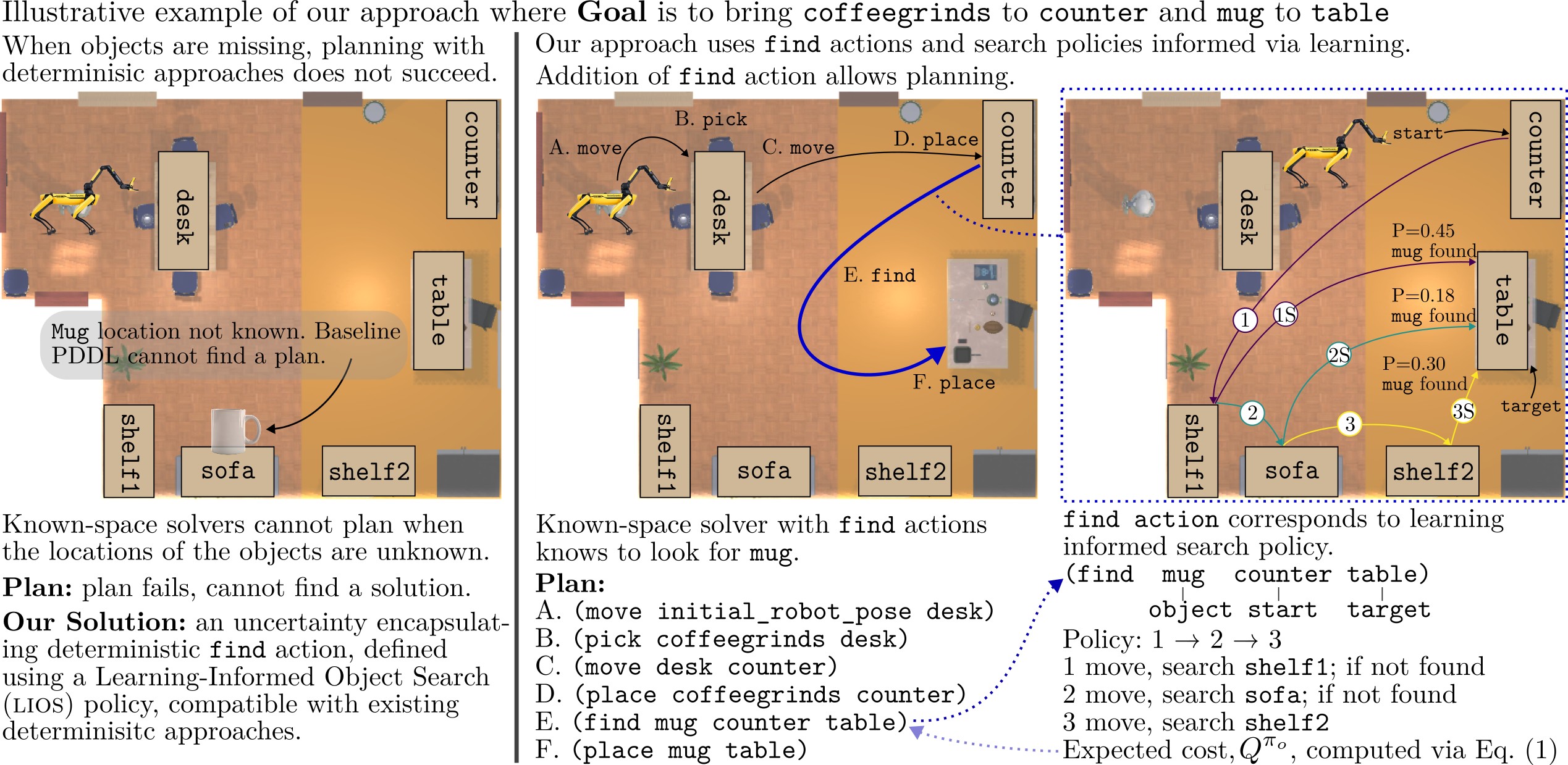}
  \vspace{-0.8em}
  \caption{\textbf{Overview: Learning-Informed Task Planning with Unknown Object Locations}. Standard task planners, like \gls{pddl} (left), typically assume full environment knowledge and cannot plan when objects are missing. Our approach (middle) adds \texttt{find}: an action to find and retrieve missing objects that high-level planning treats as deterministic. \texttt{find} actions (right) are instantiated via \emph{learning-informed object search} policies, whose expected costs inform high-level planning.}\label{fig:system-overview}
  \vspace{-1.5em}
\end{figure*}

In this work, we propose a learning-informed object search (\textsc{lios}) action that is compatible with symbolic deterministic planners, addressing all three questions above to facilitate cost-effective task planning. 
Our model predicts object location likelihoods via learning and uses model-based planning to synthesize an effective search policy, for \qnum{1}, that comes with an expected cost to find the target object. This expected cost enables the symbolic planner to decide both which object to search for \qnum{2} and when to initiate the object search \qnum{3} to minimize overall task cost.
By packaging individual object search actions inside a deterministic action, we can abstract away or \emph{encapsulate} the underlying stochasticity for high-level planning, enabling compatibility with efficient known-space planners that assume deterministic action effects.
We evaluate our approach in simulated \gls{procthor} homes on a variety of household tasks \rev{and on real-world settings using the Spot robot}. We show behaviors that interleave effective search and execution and outperform competitive learned and non-learned baselines.


\section{Related Works}
\label{sec:related-works}

\myparagraph{Task Planning}
Task planning in robotics has seen much progress over the past five decades~\cite{ghallab2016automated}. \textsc{Strips}~\cite{fikes1971strips} and \textsc{pddl}~\cite{aeronautiques1998pddl, fox2003pddl2} provide standardized languages for classical planning, yet assume full knowledge of the environment, unrealistic for robots in the wild.
\emph{Probabilistic} \gls{pddl}~\cite{younes2004ppddl1} allows for stochastic effects that could be used to model uncertainty, but such strategies struggle to scale to environments of the size and complexity needed for household-scale task planning.

\myparagraph{Object Search and Planning under Uncertainty}
The \gls{pomdp} framework~\cite{kaelbling1998} models decision-making under partial observability and probabilistic actions, making it well-suited for mobile robot planning in noisy, incomplete environments, yet scales poorly with environment size and task complexity. 
The simple \gls{lomdp} planner in Merlin et al.~\cite{merlin2024robot} handles partial observability, yet uses a greedy front-loaded object search strategy that must locate all task-relevant objects before making progress on the task.
Other recent work targets improved object search---via classical methods~\cite{veiga2016efficient, shubina2010visual, zhang2022semantic}, learning-based approaches~\cite{ye2018active, cartucho2018robust, faulhammer2016autonomous, chang2023goat, chaplot2020learning, khanna2024goat}, and foundation models~\cite{chiang2024mobility, chen2024taskclip, Zhang_2021_CVPR}---yet it is unclear how to integrate these advances into into high-level task planning.

\myparagraph{Learning and Robot Task Planning}
Recent work applies large language models (\textsc{llm}s) to task planning, using either end-to-end~\cite{pashevich2021episodic, zhang2021hierarchical} or modular~\cite{ding2023embodied, inoue2022prompter, liu2022planning, murray2022following} approaches. Many of these~\cite{gordon2018iqa, misra2017mapping, silver2022pddl, shah2023lm, wu2023embodied} assume a fully known environment. A growing direction involves generating \gls{pddl} from templates or scene descriptions, then planning with classical solvers~\cite{guan2023leveraging, liu2023llm+, silver2024generalized}.
However, these methods generally lack support for uncertainty or missing objects, and cannot reason about when or how to search during execution. Foundation model-based strategies like SayCan~\cite{ahn2022can} can operate with partial information, but often struggle on complex tasks and are not clearly compatible with classical planning.
Closer to our work, Bradley et al.~\cite{bradley2021learning} integrate learning and planning to advance \textsc{ltl}-specified tasks by revealing unseen space; their approach performs full policy search over all stochastic actions and so scales poorly with task complexity and environment scale.


\section{Problem Formulation: Task Planning with Unknown Object Locations}
\label{sec:problem-formulation}
\myparagraph{Task Planning in Fully Known Environments}
A task planning problem for a mobile robot involves finding a low-cost \emph{plan}—a sequence of actions $a_i \in A$—that transforms an initial state $s_0$ into one satisfying a goal $G$, defined by a set of logical predicates. Here, $\Oknown{}$ denotes the set of \emph{objects}, including locations or \emph{containers} (e.g., countertop, sofa) and \emph{interactable} items (e.g., mug, cellphone). Actions involve navigation or interaction (e.g., picking up a mug, activating a coffee machine), each incurring a cost such as time, distance, or effort.
The robot is assumed to have low-level skills that reliably execute movement and interaction actions in generated plans.

\myparagraph{Task Planning with Unknown Object Locations}
This work considers task planning where the location of required objects are not known. The robot knows the environment layout, container locations, and the set of available objects, yet must search in or around containers to find missing objects $\Omissing{}$.
Search actions reveal the contents 
of unsearched containers $c \in \Ounsearched{}$ and add discovered objects to $\Oknown{}$. 
Search is inherently stochastic, making the problem a challenging \emph{stochastic} \gls{mdp} incompatible with planners that presume full state knowledge. Planning seeks a policy that minimizes expected cost to complete the task.
In general, the robot lacks prior knowledge about the distribution of object locations, a gap we resolve via a learned estimator.



\section{Approach: Learning-Informed Task Planning with Missing Objects}
\label{sec:unseen-space-planning}

If we are to avoid the computational challenges of full planning under uncertainty in this domain and also maintain compatibility with popular and highly-optimized tools for robot task planning, all actions available to the robot must have deterministic postconditions, precluding straightforward inclusion of object \texttt{search} actions. 
To address this incompatibility, we introduce a novel abstraction in the form of high-level \texttt{find} actions: high-level task planning treats \texttt{find} as \emph{deterministic}---and so allows planning via languages like \gls{pddl}---yet is instantiated as an \emph{object search policy}, which includes (and abstracts away) the non-deterministic \texttt{search} actions. 
In addition to defining this abstraction, the core insights of our approach are (1) how best to define \texttt{find} and its parameters for incorporation into task planning and (2) the development of a model-based learning-informed object search strategy to instantiate \texttt{find}, which allows effective object search and comes with estimates of expected cost needed to inform high-level planning.
\texttt{find} corresponds to closed-loop search until the object is found, replanning as the environment is revealed, and so this expected cost---how long that policy is expected to take---subsumes the uncertainty of executing 
\texttt{find} action.
\begin{figure}[!t]
\begin{lstlisting}[language=Lisp,basicstyle=\fontsize{7.2}{7.2}\selectfont\ttfamily, caption={\textbf{\Gls{pddl} pseudocode for \boldtt{find}}, which serves as a \emph{deterministic} single-object search action available for high-level planning. \texttt{find} starts at location \texttt{?start} and subsequently locates and picks up the object and finally moves to location \texttt{?target}, accumulating a (learning-informed) expected \texttt{find-cost} defined via Eq.~\eqref{eq:lsp}.}, label={lst:find}]
(:action find
 :parameters (?obj - object
              ?start - location 
              ?target - location)
 :precondition (and 
   (rob-at ?start) ; robot start
   (hand-is-free)) 
 :effect (and
   (not (rob-at ?start)) (rob-at ?target)
   (not (hand-is-free)) (holding ?obj)
   ; Cost grows by (*@$Q^{\pi_o}\text{, Eq.~(1)}$@*)
   (increase (total-cost) 
      (find-cost ?obj ?start ?target))))
\end{lstlisting}

\vspace{-2.5em}
\end{figure}

\myparagraph{Defining \texttt{find} as an Uncertainty-Encapsulating High-Level Object Search Action}
We define the \texttt{find} action as monolithic single-object search skill that can be treated as having deterministic postconditions. As such, the core postcondition associated with \texttt{find} is that the object is found.
However, this condition alone still leaves considerable flexibility in how the object search occurs---search could terminate at any of the locations the object may be found---making it difficult to treat the action as deterministic.
Therefore, we encapsulate that stochasticity by defining \texttt{find} as (1) starting from a location \texttt{start}, (2) searching for \texttt{object} until it's found, then (3) picking it up and (4) moving with it to the \texttt{target} location; thus the postconditions associated with this \texttt{find} skill are that the robot is \texttt{(holding object)}, \texttt{(rob-at target)}, and if needed \texttt{(found object)}. This work treats unseen-object state variables (e.g., that a dish is \emph{clean} or \emph{dirty}) optimistically---that found objects will satisfy preconditions of later actions---and rely on replanning during deployment to repair the plan if those preconditions are not met; predicting object state variables is an opportunity for future improvement.
For \texttt{find} to be useful for informing high-level task planning, it is essential for \texttt{find(object, start, target)} to also come with an accurate \emph{expected cost} associated with its corresponding policy, so that planning may appropriately decide both \emph{what objects} the robot should look for and \emph{when it should look} for them.

\myparagraph{Learning Informed Object Search \textsc{lios}: Instantiating \boldtt{find} and Computing its Expected Cost}\label{sec:sub:expected-cost}
To instantiate the object search policy implicit in our definition of \texttt{find}, we introduce a novel \glsentryfull{LIOS} policy 
that integrates learned predictions of object locations with model-based planning to achieve both effective object search and accurate expected cost calculation.
Inspired by the planning approach of Stein et al.~\cite{pmlr-v87-stein18a} in the space of point-goal navigation in unmapped environments, we use a learned model to estimate the probability $\Pfound$ of finding a target object $o$ in each unsearched container $c \in \Ounsearched{}$.

\begin{figure}[t]
  \centering
  \vspace{.5em}
  \includegraphics[width=.49\textwidth]{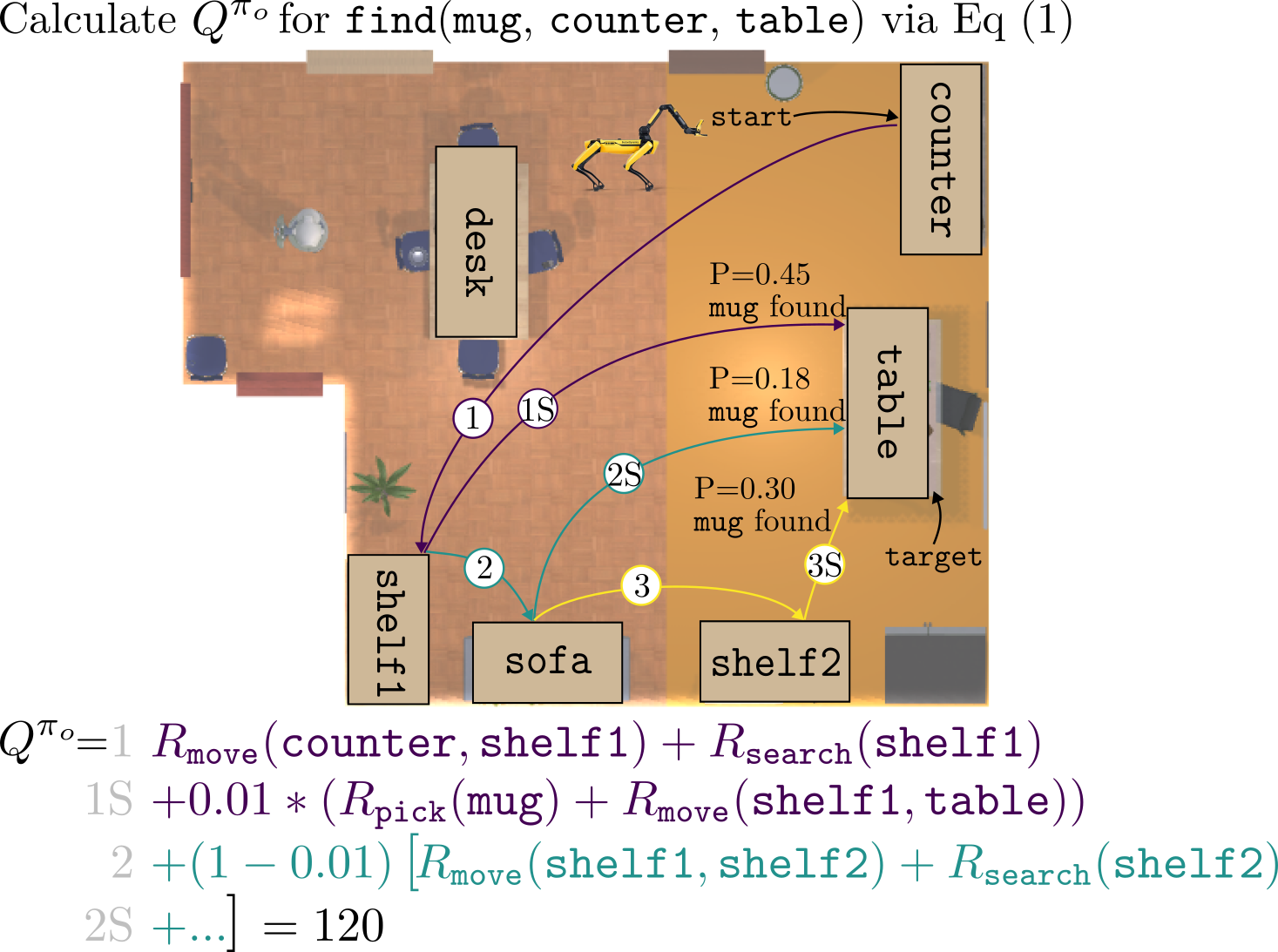}
  \vspace{-1.5em}
  \caption{Toy scenario showing computation of expected cost $Q^{\pi_o}$ for an example \gls{LIOS} policy via Eq.~\eqref{eq:lsp}. In experiments, 
  \rev{learning informs}
  $\Pfound{}$.
  }\label{fig:lsp-planning}
  \vspace{-1em}
\end{figure}

A high-level action $\texttt{find}(o, q_{\textsc{from}}, q_{\textsc{to}})$, corresponds to a search policy $\pi_o$ that starts at a location $q_{\textsc{from}}$, moves to and searches containers in a sequence until the object $o$ is found, and finally picks up that object and travels with it to $q_{\textsc{to}}$. The expected cost $Q^{\pi_o}$ of $\pi_o$ is defined via a Bellman equation:
\begin{equation}\label{eq:lsp}
\begin{split}
Q^{\pi_o}(s_t \equiv \{&q_t, q_{\textsc{to}}, \Ounsearched{} \}, a \in \Ounsearched{}) =  \\
\textit{\color{lightgray}move+search:}~& R_\texttt{move}(q_t, q(a)) + R_\texttt{search}(q(a)) \\
\textit{\color{lightgray}if object found:}~& + \Pfound(a)\left[ R_\texttt{pick}(o) + R_\texttt{move}(q(a), q_\textsc{to}) \right] \\
\textit{\color{lightgray}else not found:}~& + \left[1 - \Pfound(a)\right] Q^{\pi_o}(s_{t+1}, \pi_o(s_{t+1}))
\end{split}
\end{equation}
where $s_{t+1} \equiv \{q(a), q_{\textsc{to}}, \Ounsearched{}\!\setminus\!\{c\} \}$ updates the robot's position and marks container $c$ as \emph{searched} when the object is not found after search action $a$.
The costs $R_\texttt{move}$, $R_\texttt{search}$, etc., correspond to associated actions. Fig.~\ref{fig:lsp-planning} shows an illustrative toy example of expected cost calculation.
Planning seeks to find the policy $\pi_o$ that minimizes expected cost via Eq.~\eqref{eq:lsp}.


\myparagraph{Planning via Our Approach and Replanning as Objects are Found}
When deployed, our approach uses the partial map to synthesize the \gls{LIOS} policies and adds them to the set of available \emph{deterministic} actions, which the robot then uses to plan.
As the robot navigates, it searches containers and reveals objects, replanning as needed when new objects are found, benefiting from the new information it reveals.
To save computation in practice, \gls{LIOS} policies are recomputed only when the corresponding \texttt{find} action is selected and only eight unsearched containers are used for \gls{LIOS} policies to limit planning depth, similar to Stein et al.~\cite{pmlr-v87-stein18a}.
The robot replans after each observation until the goal is met.


\section{Experimental Setup}

\myparagraph{Household Simulation Environment: \gls{procthor}}
\label{sec:procthor}
We conduct experiments in the \gls{procthor} simulated environments~\cite{deitke2022️}, which consists of procedurally generated, realistic, large-scale household scenes. 
These environments feature diverse objects---e.g., bread, apples, coffee machines---and containers (e.g., countertops, tables, shelves), with realistic object distributions and environment layouts making them well-suited for our experiments.
To facilitate our Coffee prep task, we additionally add coffee grinds and water bottle objects to the scenes.

\myparagraph{Real-world Environment and Robot}
\label{sec:real-world}
\rev{
We conduct real-world experiments with Spot robot in home-like environment with a kitchen and a bedroom connected by a long corridor.
The environment consists of a total of nine container locations where task-relevant objects may be found and where these objects may be delivered: containers in kitchen include fridge, dining table, countertop, sink, and garbage can and containers in bedroom include bed, tv stand, desk, and couch.
}

\rev{
We use a Spot~\cite{BostonDynamicsSpot} robot from Boston Dynamics.
The robot is additionally equipped with a Spot Arm for manipulation, and a LiDAR for mapping the surroundings.
MoveIt~\cite{coleman2014MoveIt} is used for manipulator motion planning, SpotSDK~\cite{SpotSDK} is used for robot locomotion, and \texttt{ar\_track\_alvar}~\cite{ros_wiki_ar_track_alvar} package for detecting individual markers that we put on objects.
}
\begin{figure}[t]
  \centering
  \vspace{0.5em}
  \includegraphics[width=.48\textwidth]{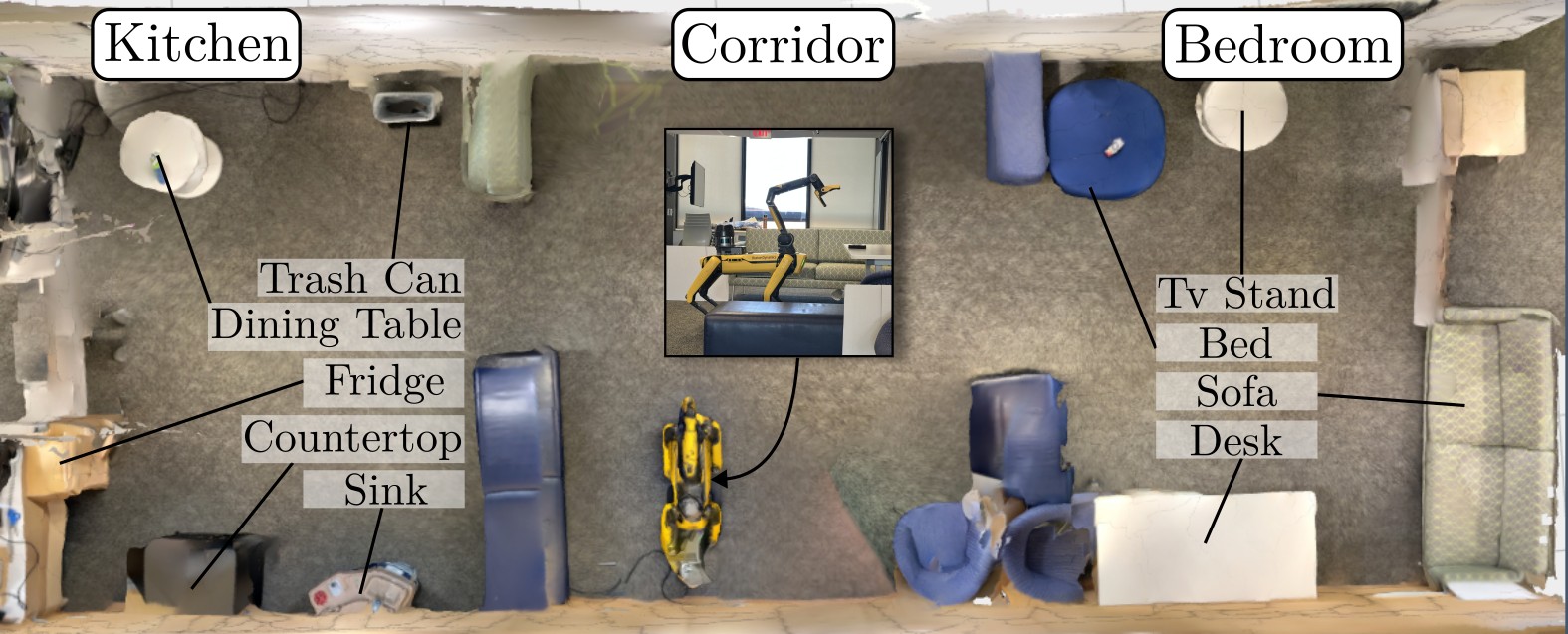}
  \vspace{-1.5em}
  \caption{A. The Boston Dynamics Spot robot, equipped with a Spot Arm for manipulation and a LiDAR sensor payload for mapping and localization. B. A top-down 3D map of the indoor home environment, showing the cluttered layout with furniture and the robot's position.} 
  \label{fig:result:realworld-setup}
  \vspace{-1.5em}
\end{figure}

\myparagraph{Action Costs and \Gls{pddl} Implementation}\label{sec:setup:pddl}
For \texttt{move} actions, cost corresponds to the travel distance between its start and end locations, computed via A* from an occupancy grid
\rev{for simulation and via \texttt{move\_base} package from waypoints for the real world.}
We use a fixed action cost of \rev{5} for all known-space operators: \texttt{pick}, \texttt{place}, \texttt{make-coffee}, \texttt{pour-water}, \texttt{peel}, \texttt{toast}.
\rev{Compared to the manipulation costs, search is less effortful and so $R_\texttt{search}$ is set to 1 on the physical robot and 0 in the simulation environments.}
We implement our approach in the popular \glsentryfull{pddl}, which we incrementally update during execution as objects and containers are revealed. We utilize the popular \textsc{ff-astar2} inadmissible search heuristic within \gls{fastdownward}, a \gls{pddl} solver.

\myparagraph{The Object Probability Estimator, Data, and Training}\label{sec:learning}
We generate training data from 500 \gls{procthor} maps~\cite{deitke2022️} to learn the probability of finding objects in each container.
For each training environment, object occurrences in containers and rooms are labeled to form a supervised learning dataset: positive labels when the object is in the target container and negative otherwise.
We extract generic names for each room, container, and target object, encode them using Sentence-\textsc{bert}~\cite{reimers-2019-sentence-bert}, and concatenate them---along with a one-hot vector indicating room, container, or object for each---into a single input vector.
We train a fully connected neural network with a cross entropy loss, which yields $\Pfound{}$ for each. 


\section{Object Search Policies and Evaluation}
\label{sec:res:object-search}
\begin{figure}
  \centering
  \vspace{.5em}
  \includegraphics[width=\columnwidth]{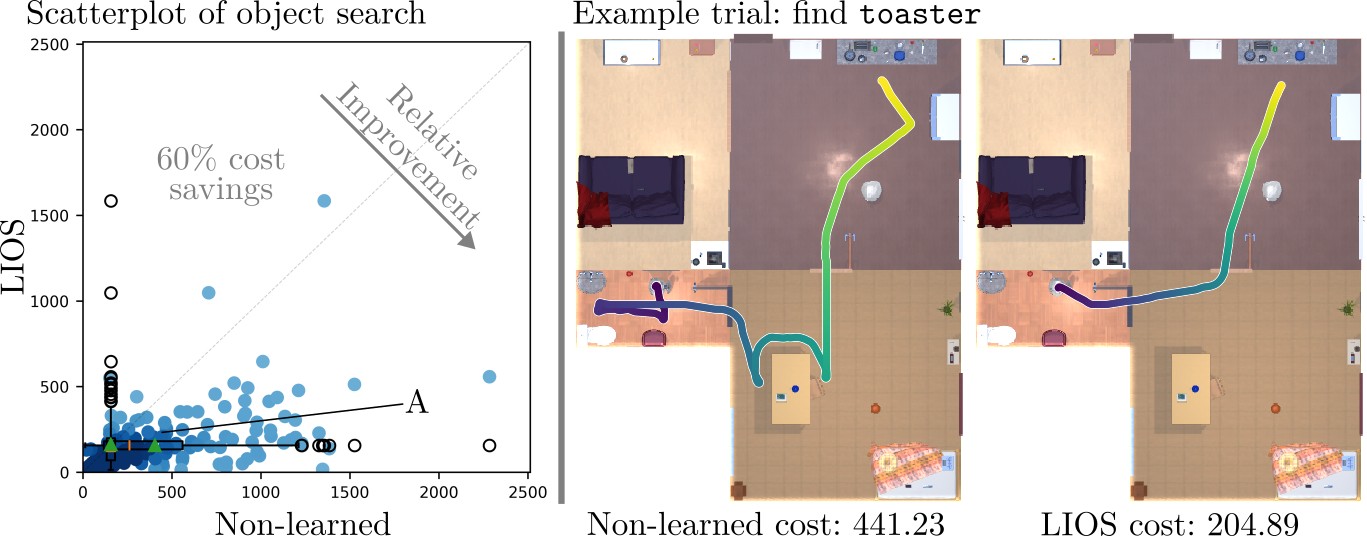}
  \vspace{-1.5em}
  \caption{\textbf{Object search performance results.} Each point in the scatter plot (left) is a single trial to find an object. Results show the improvement of our \gls{LIOS} policy versus a Greedy baseline. The example trial (right) shows trajectories from each strategy.}\label{fig:object-search}
  \vspace{-1em}
\end{figure}
We first evaluate the performance of our \gls{LIOS} policies in isolation. In each trial, the robot is placed in a previously-unseen \gls{procthor} environment and tasked to find a randomly chosen object. Across 200 trials, we compare two search strategies:
(1) \emph{Greedy}: A simple baseline strategy where the robot searches the nearest unsearched container until the object is found;
(2) \emph{\gls{LIOS}: Learning-Informed Object Search}: Our approach, which uses predicted object-location probabilities to guide search, minimizing expected cost according to Eq.~\eqref{eq:lsp}.
Each experiment executes a \texttt{find} action where the terminal location matches the starting location. After each search, the \texttt{find} process is repeated from the new position until the object is found. We record the total action cost incurred for each trial.

Our \gls{LIOS} strategy results in an improvement of 59.9\% compared to the simple Greedy baseline over the 200 object search trials.
Fig.~\ref{fig:object-search} (left) shows a scatter plot of these results in which each point represents a single trial. The plot shows that \gls{LIOS} search roughly matches or improves performance across nearly all trials.
We include a representative example trial in Fig.~\ref{fig:object-search} (right). Consistent with the statistical results, we can see that our \gls{LIOS} search, informed by predictions from our learned model, heads directly for the kitchen and so finds the toaster more quickly than the Greedy baseline.
\section{Task Planning with Missing Objects}
\subsection{Simulated Experiments and Results}

\begin{table*}[t]
\vspace{.5em}
    \caption{Avg. Cost \& Success Rate over 100 Trials for Each Planner in Each Task Scenario}
    \label{tab:result}
    \centering
    \resizebox{2\columnwidth}{!}{
    \begin{tabular}{ccc|cccccccccc}
    \toprule
    Planning  & Find & Search & \multicolumn{2}{c}{\underline{Deliver 3-Object}} & \multicolumn{2}{c}{\underline{Breakfast}} & \multicolumn{2}{c}{\underline{Coffee}} & \multicolumn{2}{c}{\underline{Breakfast+Coffee}} &  \multicolumn{2}{c}{\underline{Any-of-Three}}\\ 
    Strategies & Cost & Policy & Cost & Succ. & Cost & Succ. & Cost & Succ. & Cost & Succ.  & Cost & Succ.\\
    \midrule
    OptGreedy & Optimistic & Greedy     & 133.89 & 96\% & 118.38 & 98\% & 168.25 & 95\% & 228.36 & 90\% & 31.26 & 100\% \\ 
    PesGreedy & Pessimistic & Greedy     & 131.91 & 92\% & 114.95 & 96\%  & 152.29 & 99\% & 228.36 & 84\% & 27.85 & 100\%  \\ 
    OptLIOS & Optimistic & LIOS    & \underline{108.62} & 93\% & \underline{88.56} & 96\%  & 125.17 & 96\% & \underline{207.99} & 83\% & 23.57 & 100\%  \\ 
    PesLIOS & Pessimistic & LIOS    & 110.69 & 91\% & 94.01 & 93\%   & \underline{121.78}  & 96\% & 218.35 & 77\% & \underline{22.72} & 100\%  \\
    ModelLIOS & Model, Eq.~\eqref{eq:lsp} & LIOS  & \textbf{94.58} & 97\% & \textbf{82.59} & 97\%  & \textbf{112.76} & 99\% & \textbf{207.17} & 83\% & \textbf{22.03} & 100\%   \\
    \bottomrule
    \end{tabular}}
    \vspace{-1em}
\end{table*}

\myparagraph{Strategies for Task Planning}\label{sec:exp:policies}
To showcase the efficacy of our approach, we deploy a variety of different task planning strategies.
We contrast the performance of using our \glsentryfull{LIOS} against the non-learned Greedy object search baseline.
In addition, we also seek to show the benefit of using the expected find cost computed via our model from Eq.~\eqref{eq:lsp} over simpler baselines: an \emph{optimistic find cost baseline}---a lower bound on object search cost, as if it were guaranteed that the object would be found in the closest container, meant to encourage search---and a \emph{pessimistic find cost baseline}, the optimistic find cost plus a large penalty meant to discourage search unless absolutely necessary, similar to the aforementioned \gls{lomdp} work~\cite{merlin2024robot}. Combined, we deploy five task planning strategies that mix these approaches: two non-learned baselines \emph{OptGreedy} and \emph{PesGreedy} that rely on Greedy object search, two learned baselines \emph{OptLIOS} and \emph{PesLIOS} that use \gls{LIOS} search yet with naive estimates for the find costs, and finally \emph{ModelLIOS}: our full approach using learning-informed search and accompanying model-based find cost computed via Eq.~\eqref{eq:lsp}.

\myparagraph{Task Planning Scenarios}
We include five scenarios for task planning to showcase the performance of our approach. For all trials, if planning via \gls{pddl} does not succeed within the specified time budget, mentioned in Sec.~\ref{sec:setup:pddl}, the trial is reported as a failure and a \emph{failure cost} is returned instead. The failure cost is task-specific and chosen empirically to be greater than the maximum cost of any completed trials. For each scenario, we include the max allowable planning time $t_\text{max}$ and failure cost $R_\text{fail}$. The task planning scenarios are as follows:
\begin{tightitemize}
    \item\emph{Deliver 3-Object}\quad{}The robot is tasked to deliver three objects, chosen at random from available objects in the environment, to three random locations. The order in which the robot looks for the objects can significantly impact performance, something only our approach and its informed find cost is best-suited to address. $t_\text{max}=120 \text{ s}, R_\text{fail}=\rev{400}$
    \item\emph{Breakfast: Preparing and Serving Breakfast}\quad{}This complex task allows breakfast to be served in multiple ways: a boiled egg in a bowl, a peeled potato, tomato, or apple on a plate, or a toasted bread on a plate at a designated location. Boiling requires a pot or kettle, peeling requires a knife, and toasting requires a toaster.
    $t_\text{max}=120\text{ s}, R_\text{fail}=\rev{400}$
    \item\emph{Coffee: Preparing and Serving Coffee}\quad{}Making coffee requires coffee grinds and either, a pot, kettle, or coffee machine, that must be filled with water; water can be obtained from a water bottle. The robot must find at least 4 objects to complete the task. $t_\text{max}=240\text{ s}, R_\text{fail}=\rev{450}$
    \item\emph{Breakfast+Coffee}\quad{}The robot is tasked to \emph{both} serve breakfast and coffee, where success showcases the ability of our system to scale to complex long-horizon tasks with many possible solutions. $t_\text{max}=240\text{ s}, R_\text{fail}=\rev{450}$
    \item\emph{Any-of-Three}\quad{}The robot is tasked to retrieve any of three objects, chosen at random, and bring it back to the start location. This experiment specifically tests the efficacy of our informed expected find cost, needed to determine which object is best to seek out.
    $t_\text{max}=120\text{ s}, R_\text{fail}=\rev{100}$
\end{tightitemize}

\subsection*{Task Planning Results \& Discussion}
\label{sec:exp:results-discussion}

For our experiments, we conduct 100 trials per scenario. 
Table~\ref{tab:result} reports the success rate---accounting for failures when the \gls{pddl}-based solver cannot find a plan within the time limit---and the average per-trial cost, with failure penalties handled as described in Sec.~\ref{sec:setup:pddl}. 
In the table, the best results are shown in bold, and the second-best results are underlined.
Across all task planning scenarios, our ModelLIOS approach exhibits competitive completion rates and improved or equivalent performance compared to competitive baseline planning strategies.
Though success rates are slightly lower for the challenging many-step and open-ended Breakfast+Coffee experiments, our approach still exhibits impressive performance, emphasizing the importance of using a learning informed object search policy to quickly find missing objects.

\begin{figure*}[t]
  \centering
  \vspace{.5em}
  \includegraphics[width=\textwidth]{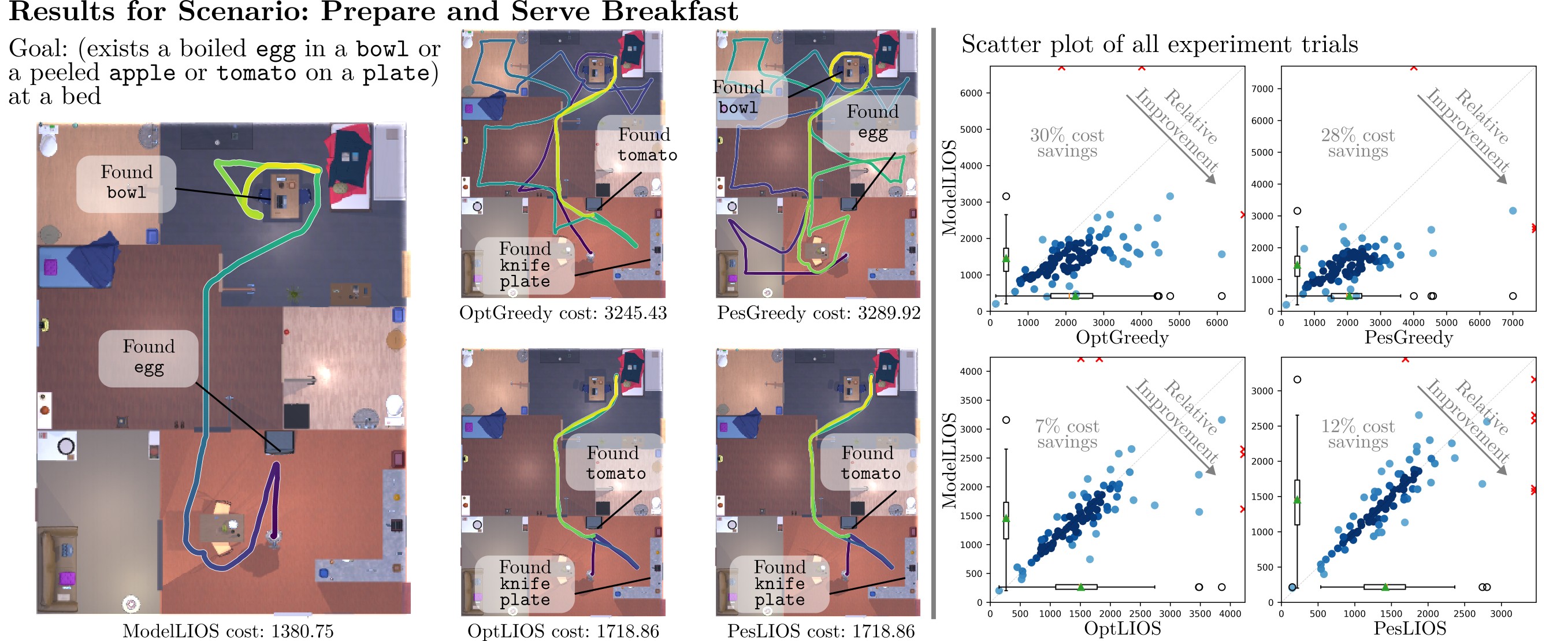}
  \vspace{-1.5em}
  \caption{\textbf{Results: Breakfast Prep Scenario}, in which the robot is tasked to prepare and serve breakfast. We include an annotated example trial (left) and scatter plots (right) that show the per-trial performance of our ModelLIOS strategy versus each baseline. Our approach effectively completes complex tasks with multiple solution pathways despite missing objects.
  }
  \label{fig:result:breakfast}
  \vspace{-.5em}
\end{figure*}

The results showcase the value of our core contributions: (i) the introduction of a high-level \texttt{find} action, (ii) using a \gls{LIOS} policy to instantiate those \texttt{find} actions for effective object search, and 
(iii) using the expected costs of those \gls{LIOS} policies as the costs for \texttt{find} to inform high-level planning.

\myparagraph{Across experiments, our results showcase the benefit of using our \gls{LIOS} strategy for a variety of household task planning scenarios.}
The planning strategies for which our \glsentryfull{LIOS} is used to instantiate the \texttt{find} actions---OptLIOS, PesLIOS, and our full ModelLIOS---are able to quickly find and retrieve objects once selected by high-level planning.
As such, the \gls{LIOS}-backed planners outperform OptGreedy and PesGreedy across all scenarios.

\begin{figure}[t]
  \centering
  \vspace{-0.5em}
  \includegraphics[width=.49\textwidth]{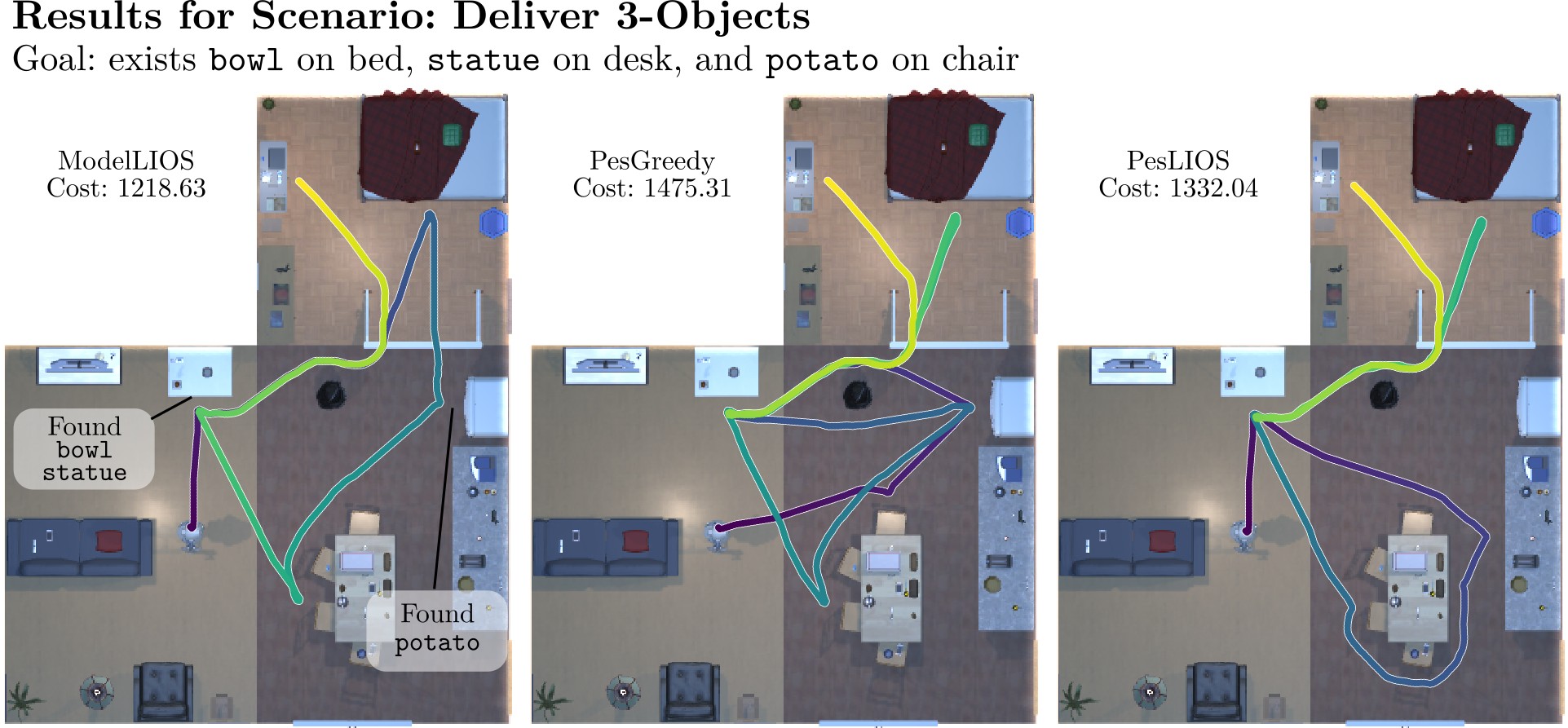}\vspace{-0.5em}
  \caption{\textbf{Deliver 3-Objects Scenario Trial}\quad{}The example trials showcase the value of both using \gls{LIOS} to quickly find objects and the importance of using its expected cost, Eq.~\eqref{eq:lsp}, to inform high-level planning, by which our ModelLIOS approach is able to determine the order in which the objects should be sought out to minimize cost.}
  \vspace{-1.5em}
  \label{fig:result:deliver3:small}
\end{figure}

\begin{figure}[t]
  \vspace{-.5em}
  \centering
  \includegraphics[width=\columnwidth]{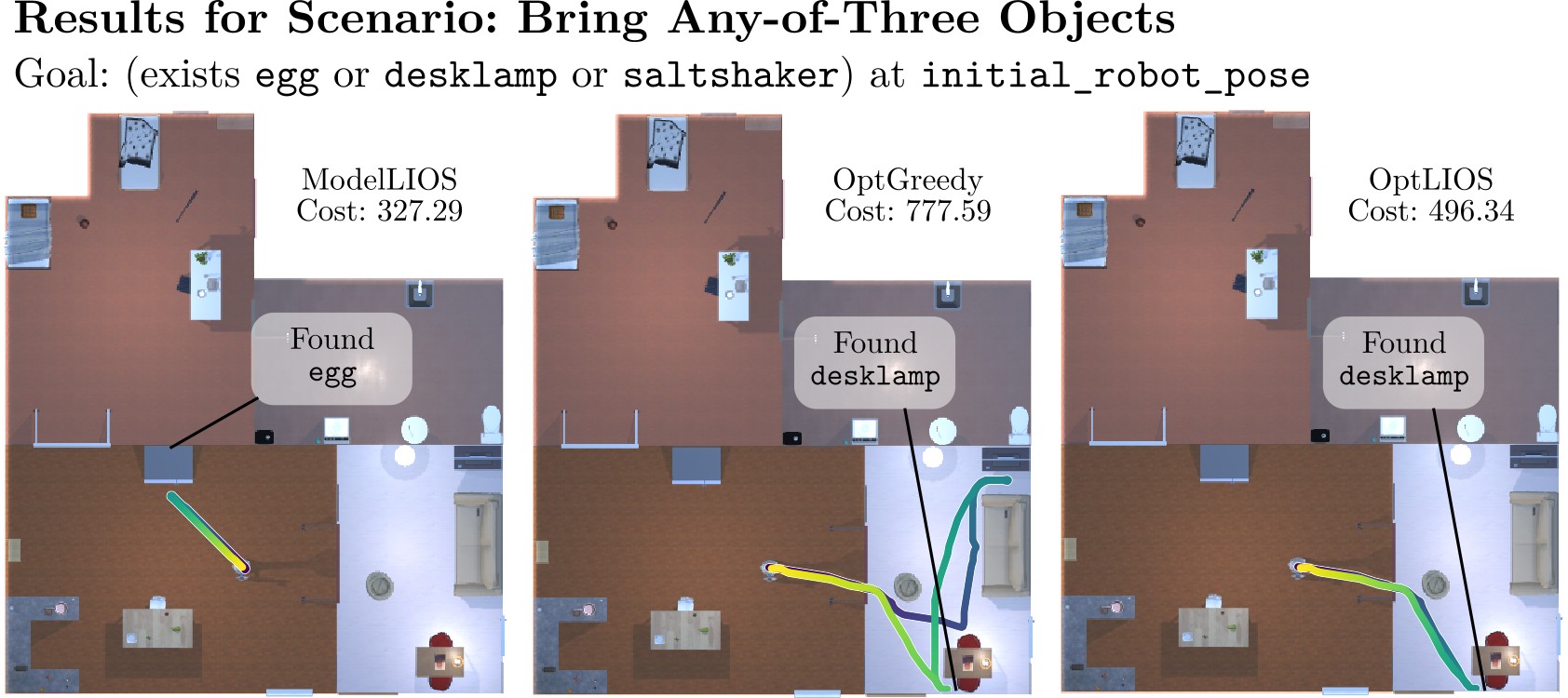}
  \vspace{-1.5em}
  \caption{\textbf{Any-of-Three Scenario Trial}, where the robot must find and bring any of three possible objects to the start.
  This scenario highlights the need for an accurate \texttt{find} cost, so that high-level planning can prioritize the easiest-to-retrieve object when given the flexibility to choose.
  }\label{fig:result:any3:small}
  \vspace{-1em}
\end{figure}

We show detailed results for the Breakfast scenario in Fig.~\ref{fig:result:breakfast}, which highlight the performance enhancements made possible by our approach.
The top scatter plots show the relative performance improvement of our ModelLIOS strategy over the Greedy baselines across the 100 trials. Moreover, the OptLIOS and PesLIOS, also able to leverage learning-informed search, are shown to improve over their Greedy counterparts. The example trials in the Breakfast (Fig.~\ref{fig:result:breakfast} left), Deliver 3-Object (Fig.~\ref{fig:result:deliver3:small}), and Any-of-Three (Fig.~\ref{fig:result:any3:small}) scenarios similarly show the benefits afforded by our learning-informed \texttt{find} actions, which are able to quickly find and retrieve the chosen objects.
Thus, the inclusion of our \texttt{find} operators and their instantiation as \gls{LIOS} policies thus serve as important components of improved behavior for task planning where object locations are not known to the robot in advance.

\myparagraph{Our results show the importance of using our expected find cost to inform high-level planning.}
Even when our \gls{LIOS} policy is used to execute the \texttt{find} actions, our results additionally show the importance of using the \gls{LIOS} expected cost, computed via Eq.~\eqref{eq:lsp}, to specify the cost associated with each find action---a necessary component for determining both \emph{what object} the robot should seek out and \emph{when to look} for each object.

We include in Fig.~\ref{fig:result:deliver3:small} a representative trial of the Deliver 3-Object scenario, comparing the performance of our ModelLIOS approach against the OptLIOS and OptGreedy strategies. While all approaches 
\rev{succeed in}
the task, only our ModelLIOS planner can both quickly seek out requisite objects and reason about what sequence of \texttt{find} actions will minimize expected cost, and so outperforms the baselines.

In the Any-of-Three scenario, the robot is tasked to locate and bring any of three candidate objects to the starting location.
Only by virtue of our learning-informed expected \texttt{find} cost computation can the robot reason about which object it should prioritize seeking out to minimize expected cost.
Leveraging this capability, our system is able to achieve the best performance among all planners. 
We highlight an example trial in Fig.~\ref{fig:result:any3:small}, which shows that our ModelLIOS is able to both determine which object is best to retrieve---via our learning-informed expected cost computed via Eq.~\eqref{eq:lsp}---and quickly retrieve it via our \gls{LIOS} policy.

Our experiments thus demonstrate the importance of incorporating an an accurate expected find cost to guide high-level planning when the robot must seek out missing objects to complete its task.

\myparagraph{Our approach affords effective task planning despite unknown object locations and is compatible with existing and highly-optimized known-space planners.}
We highlight an illustrative example trial in the Breakfast experiment of Fig.~\ref{fig:result:breakfast}. 
Despite the complexity and open-ended nature of the task---for which there are multiple possible solutions---our approach leverages both (1) our learning-informed object search strategy to quickly find requisite objects and (2) our expected \texttt{find} costs to determine which objects to prioritize and when to seek them out, resulting in improved task planning performance.

\subsection{\rev{Real-world Experiments and Results}}\label{sec:exp:real}
\begin{figure*}[t]
  \centering
  \vspace{0.5em}
  \includegraphics[width=1\textwidth]{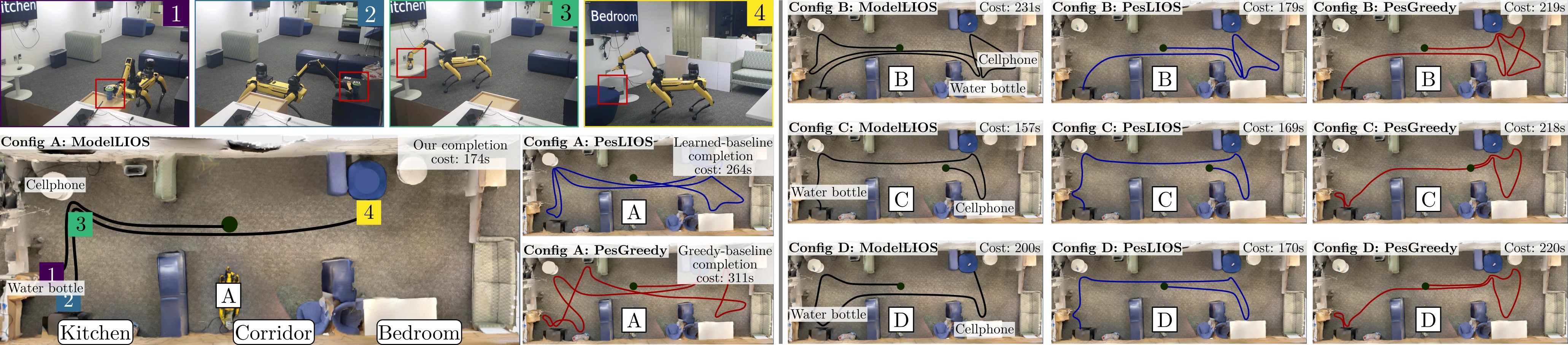}
  \vspace{-1.65em}
  \caption{\rev{
  \textbf{Real-World Results on Our Spot Platform}\quad{}We show a two-object delivery task with missing objects, where the robot has to not only reason about where to get an object but also effectively sequence when to search for which object to complete the task quickly.
  Our approach leverages both learning-informed search and computes expected costs of finding different objects to outperform both a learned baseline, which cannot effectively determine which object to look for first, and a non-learned baseline relying on greedy search for missing objects. See Sec.~\ref{sec:exp:real} for more detail.
  }}
  \vspace{-1.35em}
  \label{fig:result:realworld}
\end{figure*}

We conduct real-world demonstration with Spot robot in a home-like environment as described in \cref{sec:real-world}.
Spot is tasked to deliver the water bottle to the countertop in the kitchen and cellphone to the bed in the bedroom in any order.
We create four distinct initial environment configurations (A--D) as shown in \cref{fig:result:realworld}, wherein both task-relevant objects and the Spot robot start at different locations.
In each, we evaluate the effectiveness of our approach with respect to two baseline strategies, PesGreedy and PesLIOS. 

We highlight first the example of configuration A, in which the robot starts in the corridor, and water bottle and cellphone are on the kitchen countertop and dining table, respectively, as in \cref{fig:result:realworld}.
For both PesGreedy and PesLIOS, the robot first searches in the bedroom since their policies respectively suggest searching the bed (for PesGreedy) and desk (for PesLIOS, where it deems most effective to search for the cellphone).
After finding no relevant objects, both planners then move to the kitchen.
In contrast, our ModelLIOS, using its learning-informed model-based computation of expected cost to decide the best high-level plan, first enters the kitchen to look for the water bottle, since it reasons that finding the cellphone first is best given its initial location.

While all planners eventually complete the task,
PesGreedy searches all nine containers and PesLIOS searches six containers in order to complete the task, taking a total of 311s and 264s.
Our ModelLIOS planner outperforms both: informed by long-horizon reasoning based on model-based computation of expected costs, it requires only two container searches and completes the task in 174s total.


Performance is similar in configuration C, where the Spot robot starts near the bedroom, the water bottle is on the fridge and the cellphone is on the desk---effectively requiring the robot to move each object within the room they are located at for completing the task.
Both our ModelLIOS and the baseline PesLIOS, equipped our \gls{LIOS} policy, have similar high-level plans and outperform PesGreedy, which seeks the objects greedily to nearest locations.

Configuration D resembles C, but the robot starts nearer to the kitchen and so searches the nearby countertop, which has a high likelihood of containing the cellphone.
Though reasonable behavior, our planner slightly underperforms the PesLIOS (by 30s), a reflection of the statistical nature of planning under uncertainty.
Configuration B examines how the planners behave when objects are placed at less likely locations with respect to training data: water bottle on the desk and cellphone on the couch.
Starting at the corridor, our ModelLIOS first searches for the water bottle in the kitchen, yet incurs additional cost before turning back to search the bedroom where it finds and delivers the desired objects, slightly underperforming the baselines as a result.

\section{Limitations and Conclusion}
\label{sec:con-lim}

\subsubsection*{Limitations}
While our approach enables deterministic task planning despite unknown object locations, it has a few limitations.
First, the \texttt{find} skill focuses on searching for one object at a time within a dedicated window, lacking support for concurrent or opportunistic search during navigation. Although this simplifies planning, it may increase expected cost. 
Next, our \texttt{find} action treats object state variables optimistically and replans as necessary to repair the plan and meet preconditions for later actions; in future work, our approach could be extended to additionally predict object state variables and synthesize search policies that ensure action preconditions are satisfied by design, so as to encapsulate this additional source of uncertainty.
More broadly, our method also assumes prior knowledge of which objects are known to exist and access to a map with searchable containers---reasonable for structured domains but limiting for open-ended tasks where such information is unavailable.

\subsubsection*{Conclusion}
We present a framework for effective high-level planning when object locations are not known.
Our framework encapsulates uncertainty within \texttt{find} actions: single-object search policies that high-level planning treats as deterministic.
These actions are instantiated via effective model-based \glsentryfull{LIOS}, whose computed expected costs inform high-level task planning for effective performance. Collectively, our approach allows the robot to quickly locate requisite objects and also know what objects are best to look for and when during the plan it is best to seek them out.
Our experiments in \gls{procthor} across diverse task scenarios \rev{and in real-world using Spot robot} show that integrating our \texttt{find} actions---instantiated via \gls{LIOS} and its expected costs---into deterministic task planners enables effective task planning under partial observability.



\bibliographystyle{IEEEtran}
\bibliography{references}  

\end{document}